\documentclass{mva_style}
\usepackage{cite}
\usepackage{graphicx}
\usepackage{breqn}
\usepackage{lipsum}
\usepackage{caption}
\usepackage{subfig} 
\usepackage{color}
\usepackage{balance}
\usepackage{float}
\usepackage{multirow}

\usepackage[hyphens,spaces,obeyspaces]{url}
\urldef{\mycaseurl}\url{https://www.adaptcentre.ie/case-studies/ai-based-advert-creation-system-for-next-generation-publicity}

\finalcopy 

\begin{document}
\title{The CASE Dataset of Candidate Spaces for Advert Implantation}

\author{Soumyabrata Dev$^{1}$, Murhaf Hossari$^{1}$, Matthew Nicholson$^{1}$, Killian McCabe$^{1}$, Atul Nautiyal$^{1}$\\ 
Clare Conran$^{1}$, Jian Tang$^{3}$, Wei Xu$^{3}$, and Fran\c{c}ois Piti\'e$^{1,2}$
\thanks{Send correspondence to F.\ Piti\'e (PITIEF@tcd.ie).} 
\thanks{The  ADAPT  Centre  for  Digital  Content  Technology  is  funded  under  the  SFI Research Centres Programme (Grant 13/RC/2106) and is co-funded under the European Regional Development Fund.}
\vspace{5px} \\ 
$^{1}$The ADAPT SFI Research Centre, Trinity College Dublin\\
$^{2}$Department of Electronic \& Electrical Engineering, Trinity College Dublin\\
$^{3}$Huawei Ireland Research Center, Dublin}

\maketitle

\section*{\centering Abstract}
\textit{
With the advent of faster internet services and growth of multimedia content, we observe a massive growth in the number of online videos. The users generate these video contents at an unprecedented rate, owing to the use of smart-phones and other hand-held video capturing devices. This creates immense potential for the advertising and marketing agencies to create personalized content for the users. In this paper, we attempt to assist the video editors to generate augmented video content, by proposing candidate spaces in video frames. We propose and release a large-scale dataset of outdoor scenes, along with manually annotated maps for candidate spaces. We also benchmark several deep-learning based semantic segmentation algorithms on this proposed dataset.
}

\section{Introduction}

Impressive technological advances have provided a platform to create and share on-demand video contents over the internet. The consumers have an insatiable appetite for media, and artificial intelligence (AI)-based techniques assist us in understanding such media~\cite{hossari2018test}. This creates opportunity for the marketing and the advertisement agencies to generate personalized marketing products, according to the consumers' likes and dislikes. However, the current \emph{skip-ad} generation have a limited attention span~\cite{chalawsky2013ad}, and therefore, new strategies of advertisement needs to be employed. One such technique include product placement or embedded marketing, that involves the video editors to artificially integrate new adverts into original scenes.

\begin{figure}[htb]
\centering
\subfloat[Floating candidate]{\includegraphics[height=0.14\textwidth]{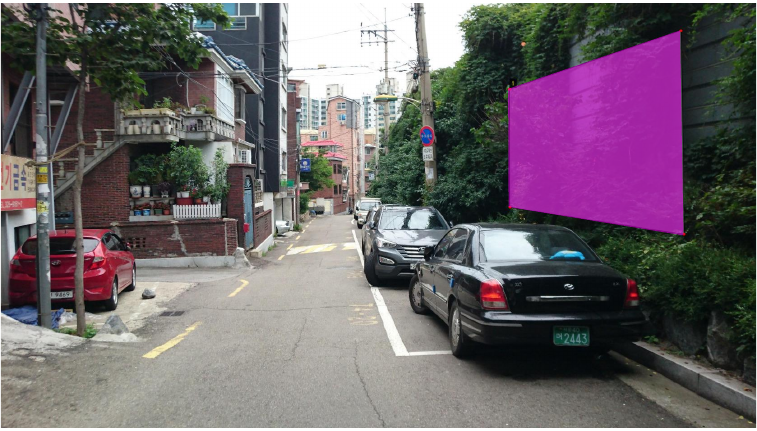}}\ 
\subfloat[Anchored candidate]{\includegraphics[height=0.14\textwidth]{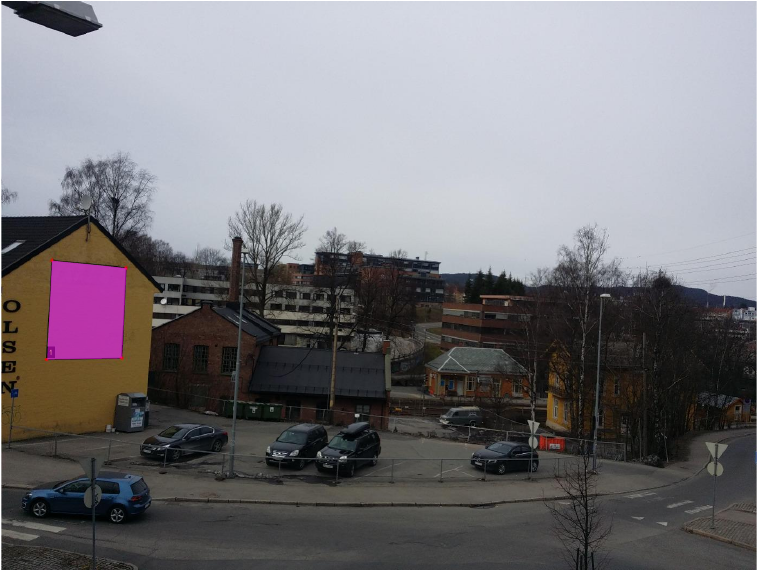}}\ 
\caption{Potential spaces for advertisement integration in outdoor scenes. The candidate spaces can either be floating across the scene, or anchored against walls in the scene.}
\label{fig:wall-story}
\vspace{-0.4cm}
\end{figure}

Online videos provide several opportunities for integrating new advertisements in the original videos. We developed a deep-learning based advert creation system~\cite{nautiyal2018advert} that can automatically identify the frames in a video containing an advertisement~\cite{hossari2018adnet}. Additionally, we also identify the specific frames in a video where new advertisements can be artificially augmented, conforming to the subjective human judgement. 
It is important to select the candidate space in a frame in a systematic manner, such that the new advert can be augmented in a seamless and sensical manner. For example, adding advertisement boards by the side of the road, and onto empty wall spaces are good candidates for advert integration. Figure~\ref{fig:wall-story} describes such a scenario, wherein candidate spaces are marked based on the scene understanding. These candidate spaces of new advertisements in a video frame are generally manually selected by the video-editors during the post-processing stage.

In most cases, the video editors uses his/her understanding of the scene, and proposes a new space for advert integration. In the literature, there are no related works that can automatically parse the scene in the image, and propose a candidate space for integrating advertisements. We attempt to bridge this gap in the literature, by releasing the first large-scale dataset of outdoor scenes, with manually annotated candidate spaces. 

The main contributions of this paper are as follows: (a) we propose and release the first large scale dataset of candidate placements in outdoor scene images, (b) we also provide a systematic and detailed benchmarking results of several popular deep learning based segmentation frameworks in our proposed dataset. We believe that this dataset will greatly serve the multimedia and advertisement communities, and assist in providing interesting research in this field. The rest of the paper is arranged as follows: Section~\ref{sec:data} describes the dataset, and its characteristics. We also provide a few benchmarking results on the proposed dataset in Section~\ref{sec:exp}. Finally, Section~\ref{sec:conc} concludes the paper, and describes our future work.

\section{Dataset}
\label{sec:data}
We refer our dataset as \textbf{CASE} dataset, that stands for \textbf{CA}ndidate \textbf{S}paces for adv\textbf{E}rt implantation dataset~\footnote{The download link of the dataset is available here: \mycaseurl}. The sources of the images in this dataset is Cityscapes dataset~\cite{cordts2016cityscapes} comprising street view images. We randomly select $10000$ images from the Cityscapes dataset that for annotating with potential placements for advertisements.

\begin{figure*}[]
\centering

\includegraphics[height=0.11\textwidth]{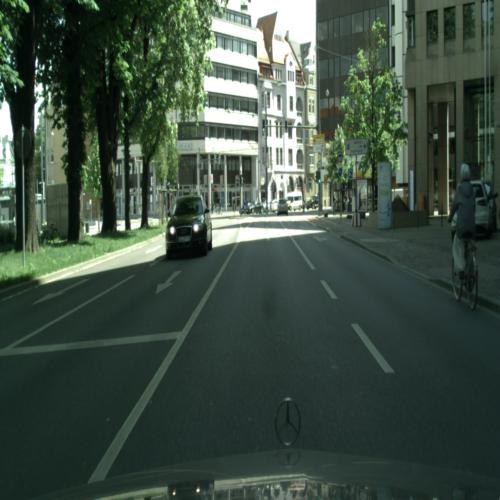}
\includegraphics[height=0.11\textwidth]{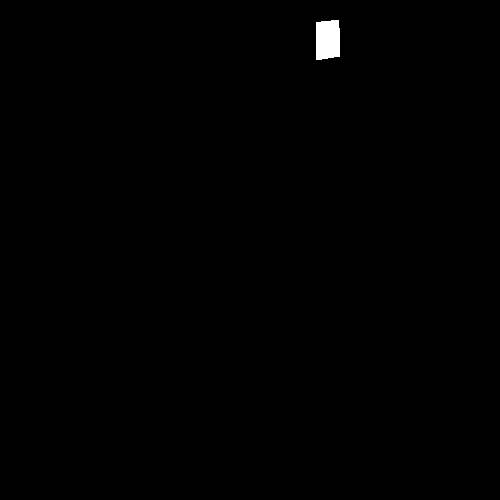}
\includegraphics[height=0.11\textwidth]{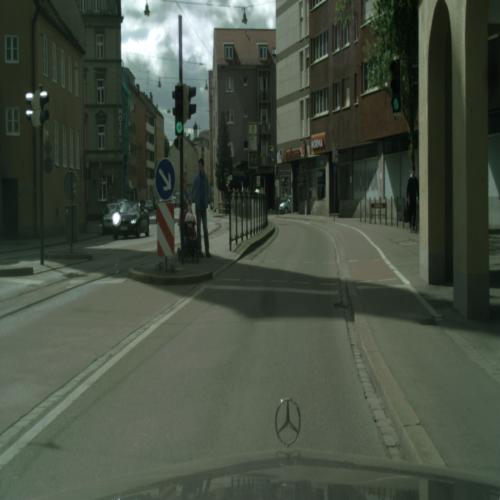}
\includegraphics[height=0.11\textwidth]{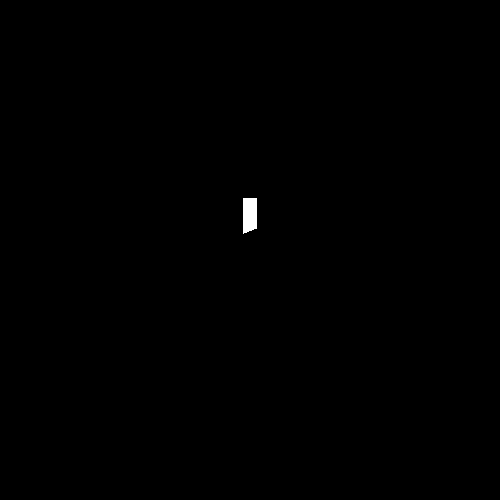}
\includegraphics[height=0.11\textwidth]{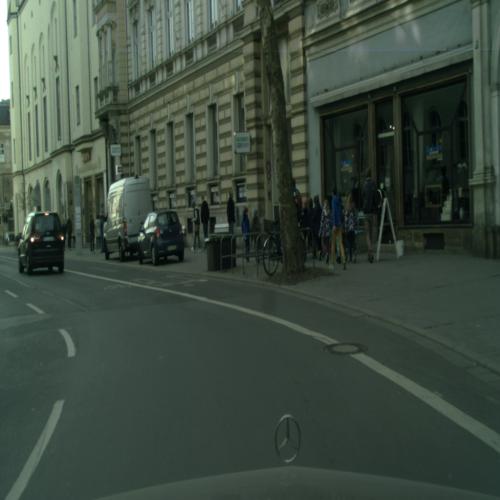}
\includegraphics[height=0.11\textwidth]{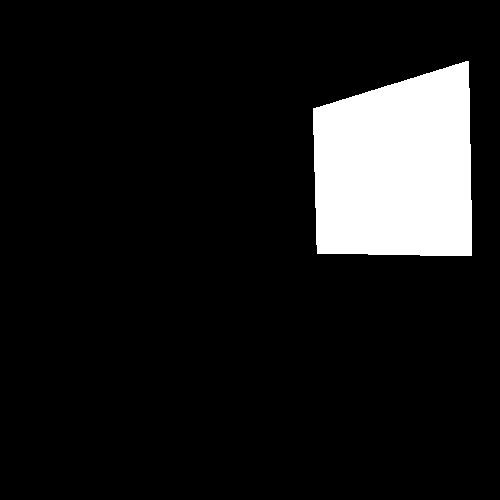}
\includegraphics[height=0.11\textwidth]{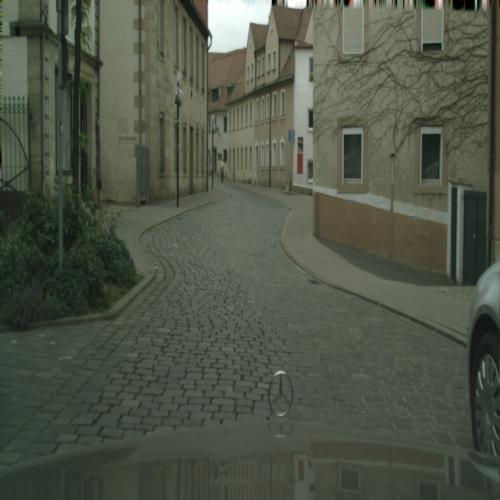}
\includegraphics[height=0.11\textwidth]{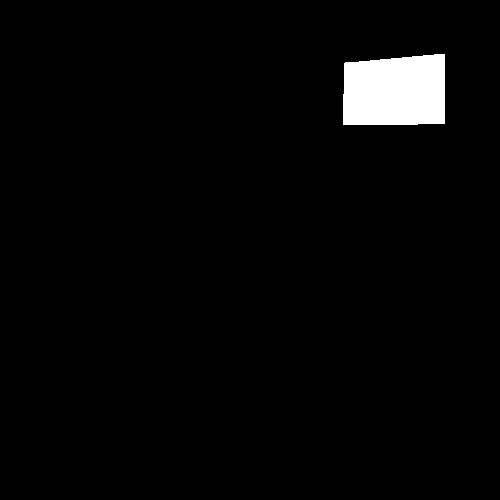}\\
\includegraphics[height=0.11\textwidth]{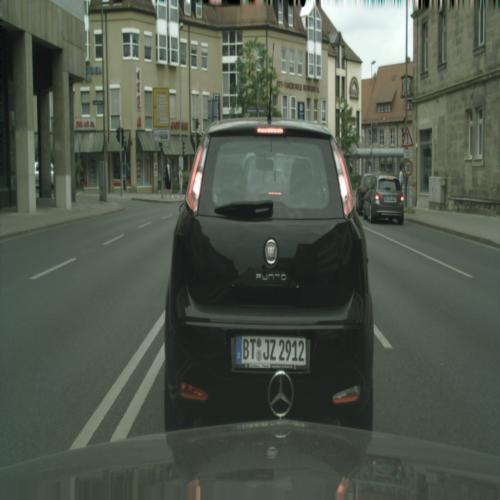}
\includegraphics[height=0.11\textwidth]{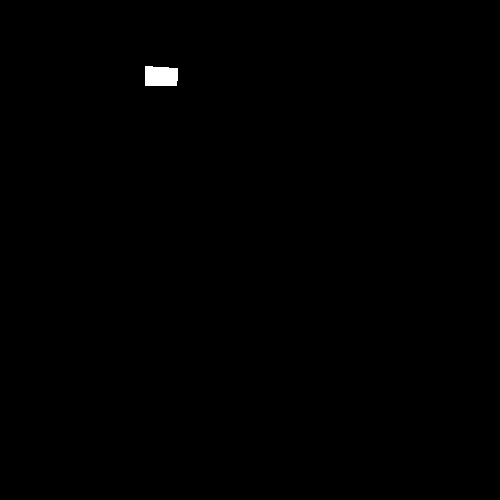}
\includegraphics[height=0.11\textwidth]{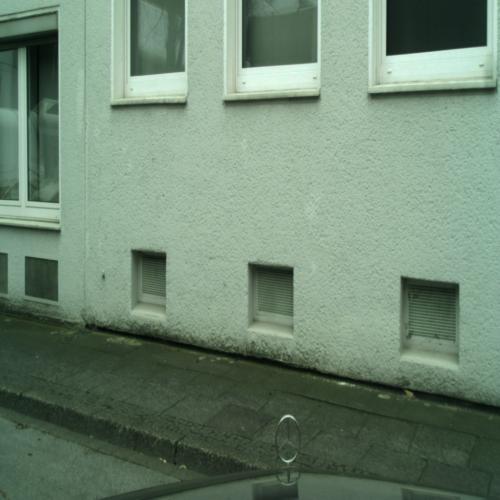}
\includegraphics[height=0.11\textwidth]{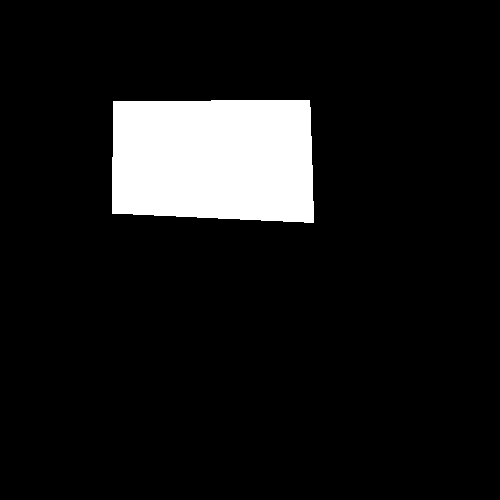}
\includegraphics[height=0.11\textwidth]{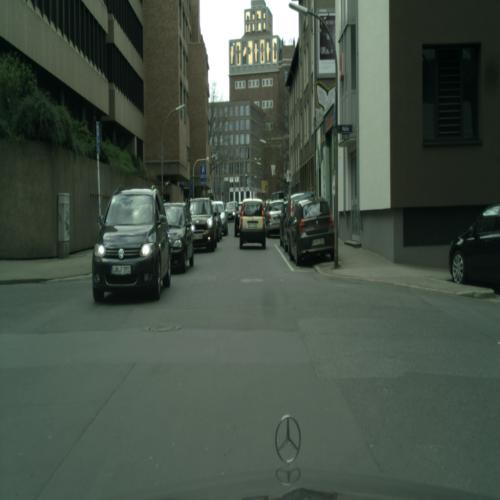}
\includegraphics[height=0.11\textwidth]{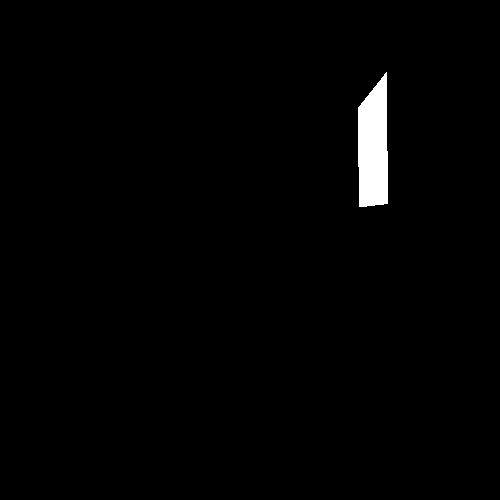}
\includegraphics[height=0.11\textwidth]{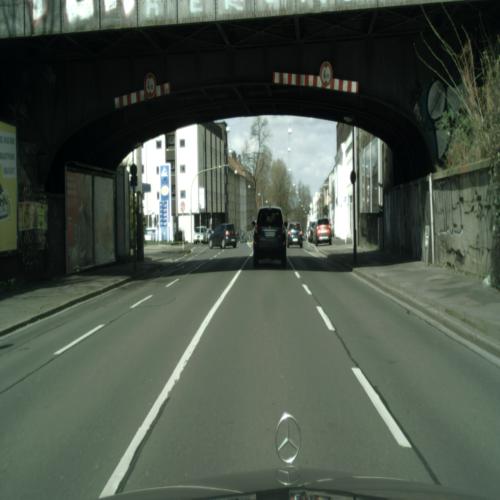}
\includegraphics[height=0.11\textwidth]{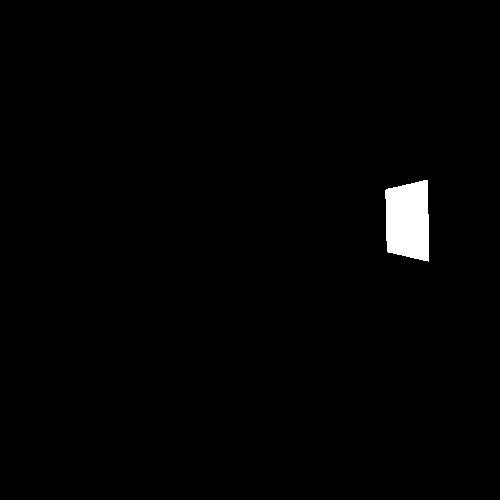}\\
\includegraphics[height=0.11\textwidth]{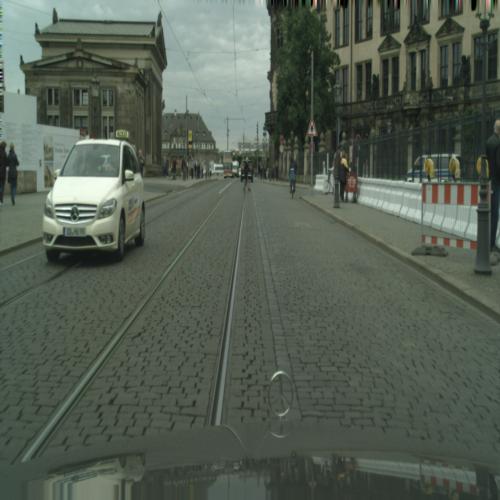}
\includegraphics[height=0.11\textwidth]{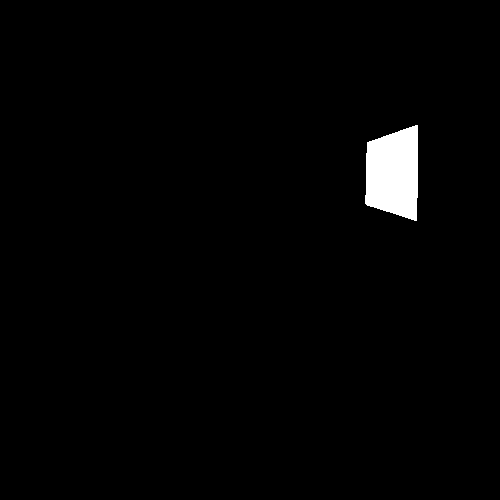}
\includegraphics[height=0.11\textwidth]{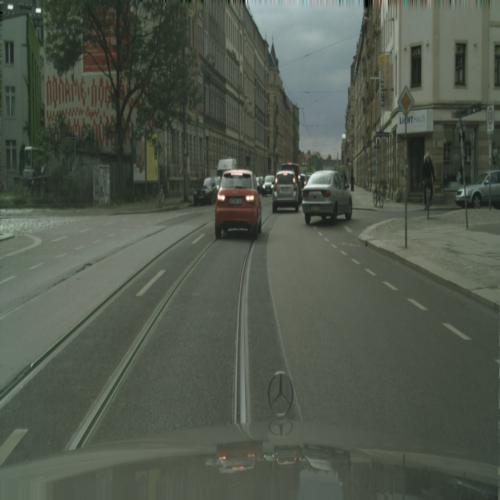}
\includegraphics[height=0.11\textwidth]{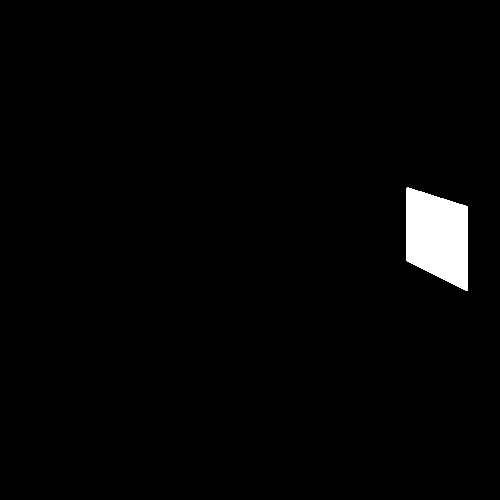}
\includegraphics[height=0.11\textwidth]{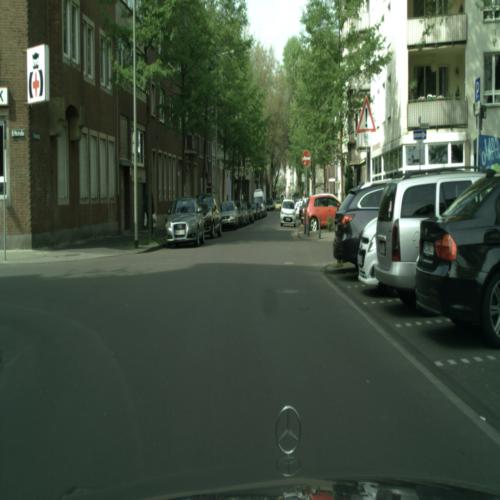}
\includegraphics[height=0.11\textwidth]{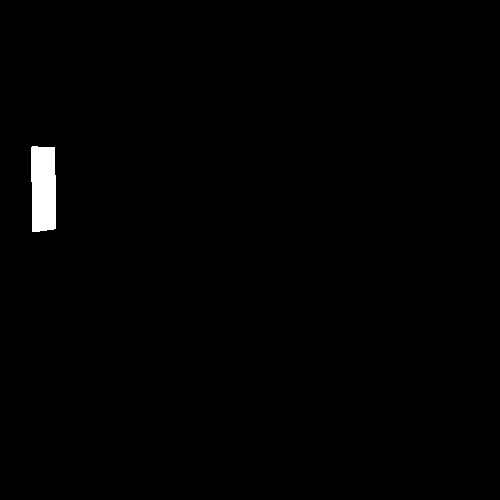}
\includegraphics[height=0.11\textwidth]{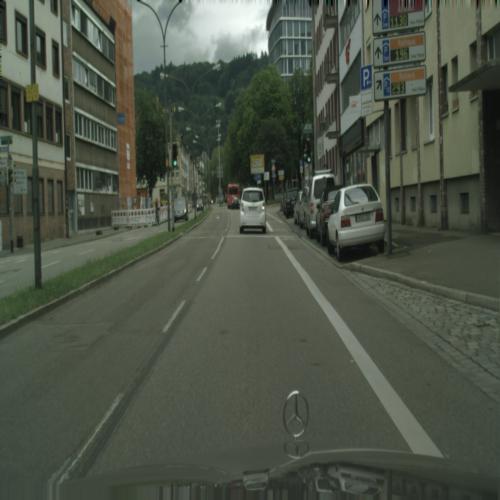}
\includegraphics[height=0.11\textwidth]{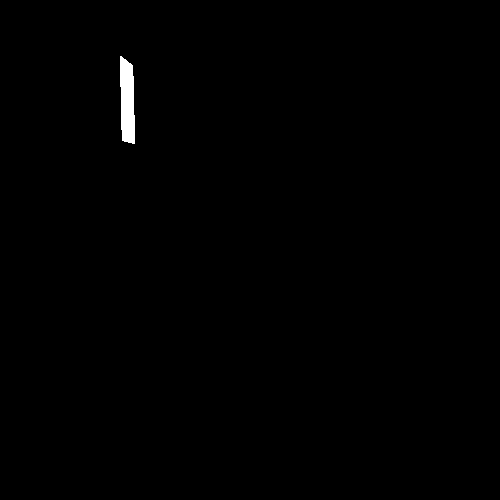}\\
\includegraphics[height=0.11\textwidth]{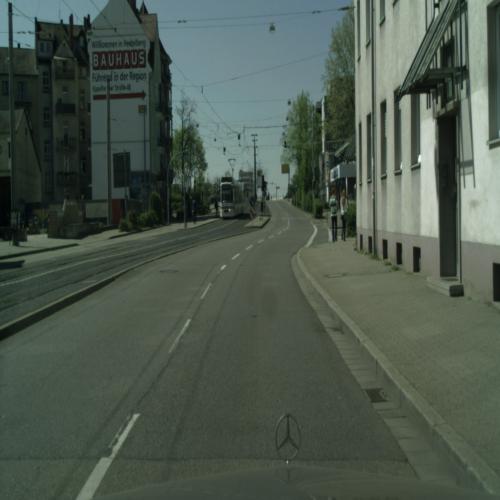}
\includegraphics[height=0.11\textwidth]{mask3183.png}
\includegraphics[height=0.11\textwidth]{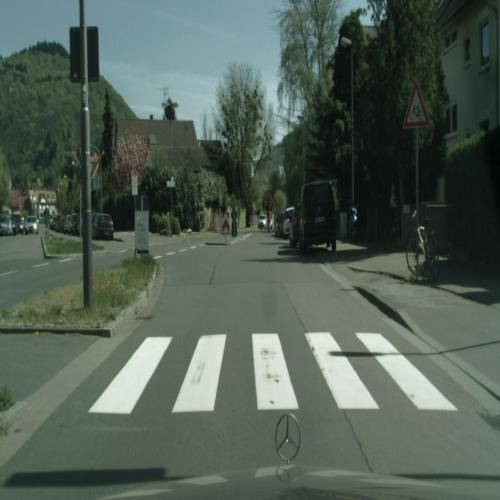}
\includegraphics[height=0.11\textwidth]{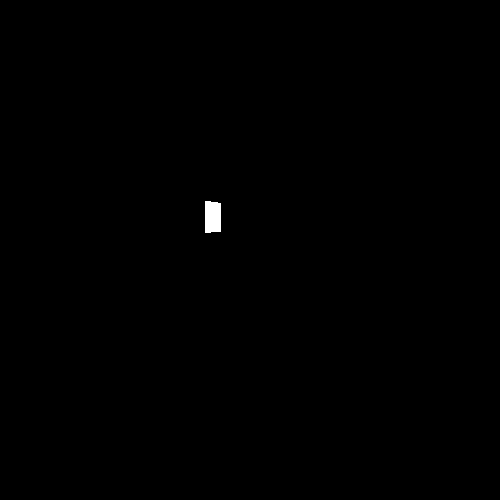}
\includegraphics[height=0.11\textwidth]{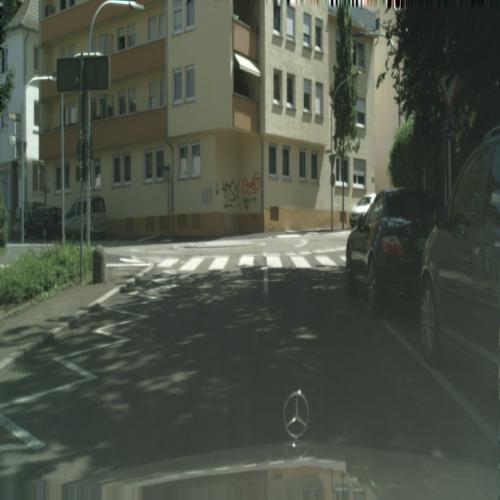}
\includegraphics[height=0.11\textwidth]{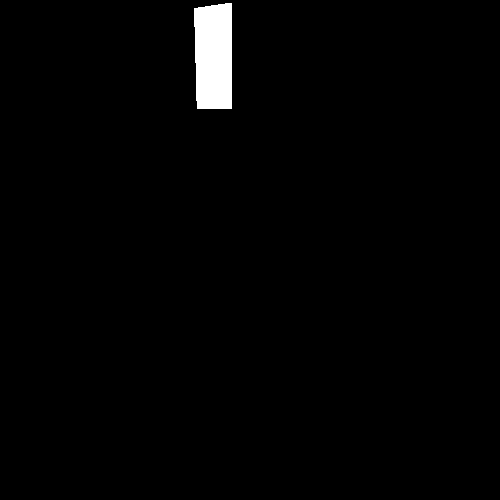}
\includegraphics[height=0.11\textwidth]{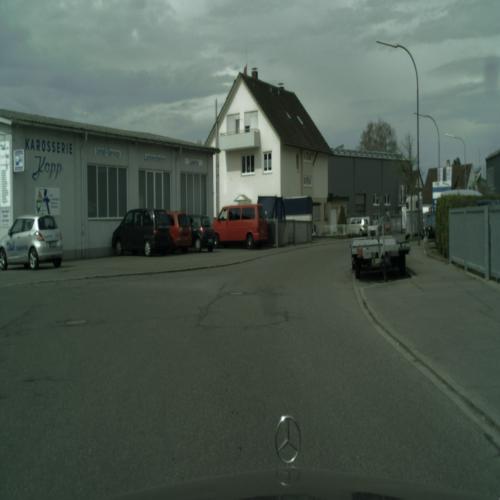}
\includegraphics[height=0.11\textwidth]{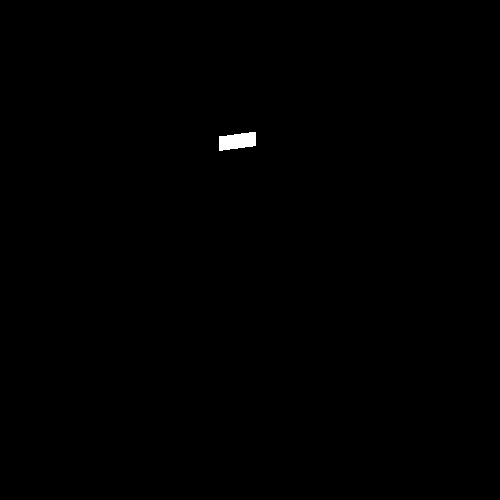}
\caption{Representative images of the CASE dataset, along with manually annotated binary candidate maps for integrating new adverts.}
\label{fig:sample-images}
\vspace{-0.6cm}
\end{figure*}

\subsection{Creation of candidate spaces}
We are interested in learning candidate placement of objects within a scenes. We have chosen to investigate placing regularly shaped billboard in street view images. In our application case study, billboards are considered as rectangular areas that can be added to the sides of buildings, or fixed by the pathways and road sides for placement of posters or advertisements. We employ paid volunteers to manually go through each of our selected $10000$ images, and mark the location where they can assume that a billboard could be placed. This process is done carefully, such that the embedded advertisement matches the perspective of the scene. We use the VGG Image Annotator tool~\footnote{The tool is available via \url{http://www.robots.ox.ac.uk/~vgg/software/via/via.html}.} for annotating the candidate spaces. The user annotate each image in the dataset by selecting $4$ points, that either form an anchored placement (against wall or surface) or floating placement in the scene.

Although there can be several potential spaces to embed the new advertisement, we restrict the annotators to include only a single candidate placement in an image. This is intentionally done, such that the annotator chooses the \emph{best} candidate space for integration, according to his/her subjective judgment. Examples of good placement include, but not limited to, placing advert on a blank wall, beside existing billboards, hanging beside a lamppost, or over a window/group of windows.  We do not annotate existing billboards or posters. This process is completed for all the selected images of our dataset. Post the annotation process, we manually go through all the generated candidate spaces, to ensure that the annotations are performed properly.

Figure~\ref{fig:sample-images} shows a few representative images of the CASE dataset, along with its manually annotated binary masks. The binary masks are generated from the four corners of the annotated billboard.

\subsection{Dataset characteristics}
\begin{figure}[htb]
\centering
\includegraphics[width=0.41\textwidth]{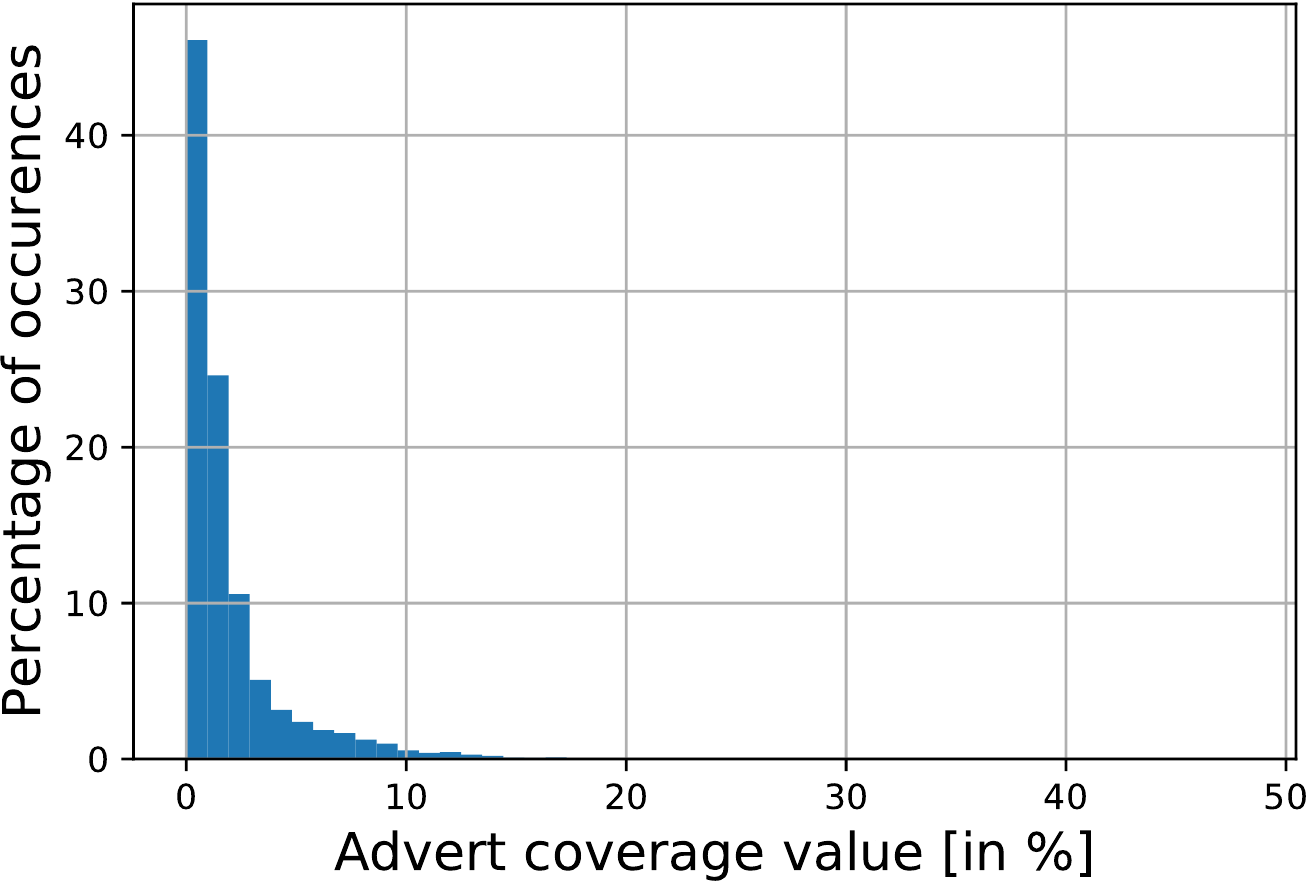}
\caption{Statistical distribution of the percentage of occurrences of the advertisement spaces w.r.t. the advert coverage values.}
\label{fig:wall-coverage}
\vspace{-0.5cm}
\end{figure}

The candidate spaces in the images are annotated with varying sizes. We define the amount of image area covered with a candidate space as the advert coverage value. We ensure that the CASE dataset has a varying values of advert coverage values, across all the images. Figure~\ref{fig:wall-coverage} describes the distribution of the coverage of the candidate spaces. 

\section{Benchmarking Experiments}
\label{sec:exp}

In addition to the proposed dataset of candidate spaces, we also provide a few benchmarking experiments of several semantic segmentation algorithms in the CASE dataset. With the success of deep neural networks, several convolutional neural networks have been proposed that provides impressive results on various semantic segmentation tasks. We benchmark the performance of fully convolutional neural network (FCN)~\cite{long2015fully}, pyramid scene parsing network (PSPNet)~\cite{zhao2017pyramid}, and U-Net~\cite{ronneberger2015u}.

The FCN network by Long et al.\ employs only locally connected layers, without the use of any dense layers in its architecture. The layers in FCN include convolution, pooling and upsampling layers. It provides dense segmentation masks of input images of any dimension. Zhao et al.\ in ~\cite{zhao2017pyramid} introduced a pyramid pooling module, that can better grasp the context of the input image. The PSPNet produced impressive results on the ImageNet scene parsing challenge~\cite{zhou2016semantic}. The advantage of the pyramid pooling module is that it captures information at varying scales, across different sub-regions of the input image. Recently, Ronneberger et al.\ introduced a convolutional neural network called U-Net that provided impressive results in biomedical image segmentation. The U-Net network has a symmetric structure, and consists of three primary parts -- contracting, bottleneck, and expanding path~\cite{ronneberger2015u}. It provides skip connection between the contracting and expanding paths, that provides the local information to the global information, during the upsampling process. These networks are popularly used in the area of semantic segmentation, for an efficient generation of segmentation masks. We use these networks as benchmarking models for our proposed dataset. Currently, there are no networks that are specifically designed for identifying candidate spaces in input images.

\begin{figure*}[htb]
\centering
\includegraphics[height=0.18\textwidth]{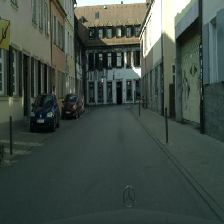}
\includegraphics[height=0.18\textwidth]{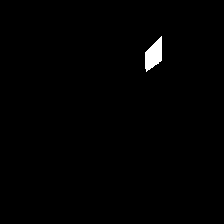}
\includegraphics[height=0.18\textwidth]{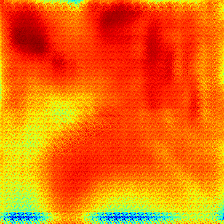}
\includegraphics[height=0.18\textwidth]{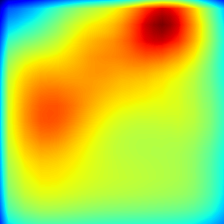}
\includegraphics[height=0.18\textwidth]{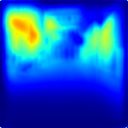}\\
\includegraphics[height=0.18\textwidth]{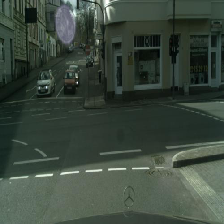}
\includegraphics[height=0.18\textwidth]{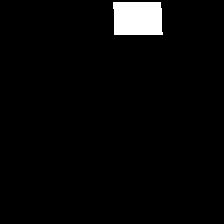}
\includegraphics[height=0.18\textwidth]{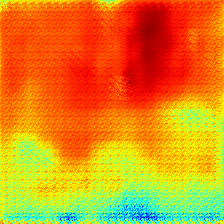}
\includegraphics[height=0.18\textwidth]{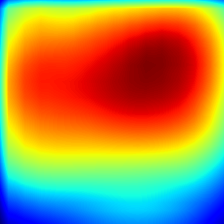}
\includegraphics[height=0.18\textwidth]{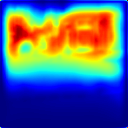}\\
\includegraphics[height=0.18\textwidth]{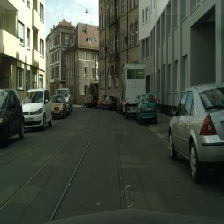}
\includegraphics[height=0.18\textwidth]{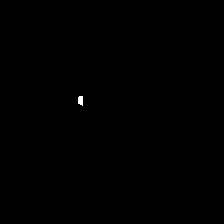}
\includegraphics[height=0.18\textwidth]{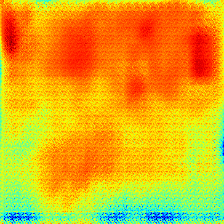}
\includegraphics[height=0.18\textwidth]{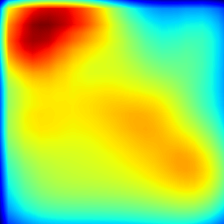}
\includegraphics[height=0.18\textwidth]{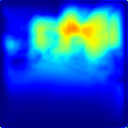}\\
\includegraphics[height=0.18\textwidth]{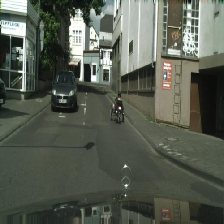}
\includegraphics[height=0.18\textwidth]{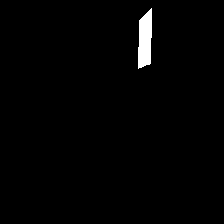}
\includegraphics[height=0.18\textwidth]{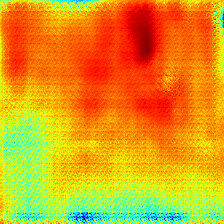}
\includegraphics[height=0.18\textwidth]{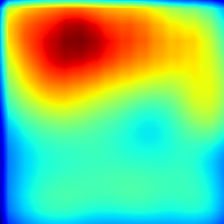}
\includegraphics[height=0.18\textwidth]{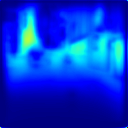}\\
\vspace{-0.3cm}
\subfloat[Input image]{\includegraphics[height=0.18\textwidth]{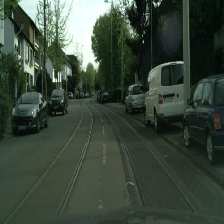}}\ 
\subfloat[Ground truth]{\includegraphics[height=0.18\textwidth]{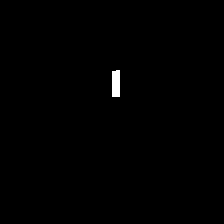}}\ 
\subfloat[FCN result]{\includegraphics[height=0.18\textwidth]{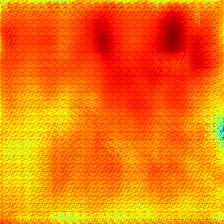}}\ 
\subfloat[PSPNet result]{\includegraphics[height=0.18\textwidth]{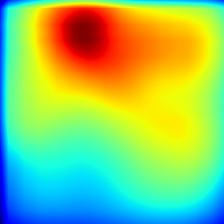}}\ 
\subfloat[U-Net result]{\includegraphics[height=0.18\textwidth]{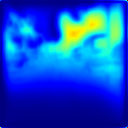}}
\caption{Subjective evaluation of various segmentation algorithms on CASE dataset. The results in (c--e) are visualized, prior to the final \emph{softmax} layer of the individual networks. The probabilistic maps of FCN are inverted, in order to illustrate the detection of candidate space (instead of non-candidate space) class.}
\label{fig:subj-eval}
\vspace{-0.7cm}
\end{figure*}

\subsection{Subjective Evaluation}

Figure~\ref{fig:subj-eval} shows a few sample results of the benchmarking algorithms on the CASE dataset. We visualize the results obtained from these algorithms, prior to the final \emph{softmax} layer. This is useful, as it provides us a probabilistic viewpoint on the possibility of a candidate space in an image. 
Most of these approaches generate a lot of false positives in the generated binary maps. This is understandable because there is no \emph{unique} position of advertisement in an input image. The position of the embedded advertisement is completely dependent on the subjective judgement of the annotator. It is interesting to observe from Fig.~\ref{fig:subj-eval}(d) and (e), that PSPNet and U-Net identifies the buildings and the sideways of the roads as possible placements for new adverts. The FCN network, on the other hand is conservative in nature -- it produces significantly fewer false positives, and do not propose a particular area of the image as a candidate space for advertisement integration. 
However, we observe from the visual results, that most of these networks viz.\ PSPNet and U-Net can successfully learn the semantics of the input scenes, and provide reasonable maps of candidate spaces.

\subsection{Objective Evaluation}
In addition to the subjective evaluation of the benchmarking algorithms, we also provide an objective evaluation of the different approaches. We report the average values of the following metrics -- pixel accuracy, mean accuracy, mean intersection over union, and frequency weighted intersection over union. These metrics are variations of classification accuracy and intersection over union (IOU). Let us assume that the total number of pixels belonging to class $i$, and predicted to belong to class $j$, is $n_{ij}$. The total number of classes in this particular task of semantic segmentation is assumed to be $n_{cl}$.

\begin{equation}
\mbox{Pixel Accuracy} = \frac{\sum_{i}^{} n_{ii}}{\sum_{i}^{} t_{i}}
\end{equation}

\begin{equation}
\mbox{Mean Accuracy} = \frac{1}{n_{cl}}\sum_{i}^{}\frac{n_{ii}}{t_i}
\end{equation}

\vspace{-0.3cm}

\begin{dmath}
\mbox{Mean } \mbox{Intersection } \mbox{Over } \mbox{Union} = \frac{1}{n_{cl}}\frac{\sum_{i}^{}n_{ii}}{t_i+\sum_{j}^{}n_{ji}-n_{ii}}
\end{dmath}

\vspace{-0.7cm}

\begin{dmath}
\mbox{Frequency } \mbox{Weighted }\mbox{Intersection } \mbox{Over } \mbox{Union} = \frac{1}{\sum_{k}^{}t_k}\frac{\sum_{i}^{}t_in_{ii}}{t_i+\sum_{j}^{}n_{ji}-n_{ii}}, 
\end{dmath}

where $t_i = \sum_{j=1}^{n_{cl}} n_{ij}$ is the total number of pixels in class $i$.

Table~\ref{table:result} summarizes the results of the various benchmarking algorithms in CASE dataset. 

\begin{table}[ht]
\begin{tabular}{l|p{1.3cm}|p{1.3cm}|p{1.3cm}|p{1.3cm}}
       & Pixel Accuracy & Mean Accuracy & Mean IOU & Frequency Weighted IOU \\ \hline
FCN    & \textbf{0.978} & 0.509 &  \textbf{0.498}  &  \textbf{0.959}     \\ \hline
PSPNet & 0.545 & 0.625 & 0.284 & 0.529 \\ \hline
U-Net & 0.619 & \textbf{0.727} &   0.327 &  0.601        \\ 
\end{tabular}
\caption{Benchmarking of the CASE dataset with various deep-learning based segmentation algorithms. The best performance for each metric is indicated in bold.}
\label{table:result}
\end{table}

Amongst the benchmarking networks, we observe that the FCN network has the best evaluation scores. The FCN network can learn the semantics of the annotated scene, and provides a good understanding of a candidate space in an image. The results obtained from PSPNet and U-Net can be further improved by training these networks on larger dataset of diverse image types, and employing batch normalization across multiple graphics processing units (GPUs). These results provide a detailed benchmark of the popular segmentation algorithms on the candidate space dataset. The results can be further improved, by designing a bespoke neural network that is specifically designed for the task of identifying candidate spaces in outdoor scenes. 

\section{Conclusion and Future Work}
\vspace{-0.3cm}
\label{sec:conc}
In this paper, we propose CASE -- a large-scale dataset of outdoor scenes with manually annotated binary maps of candidate spaces. This dataset is the first of its kind that assists the video editors an efficient and systematic manner of creating augmented video contents with new advertisements. In the future, we intend to relax this criterion of outdoor scenes, and also further extend this dataset, by including images from indoor scenes and entertainment television shows.

\vspace{-0.3cm}
\section*{Acknowledgement}
\vspace{-0.3cm}
The authors would like to acknowledge the contribution of the various paid volunteers, who assisted in the creation of high-quality candidate maps in this dataset. 


\bibliographystyle{splncs04}

\end{document}